\documentclass[runningheads]{llncs}
\usepackage[T1]{fontenc}

\usepackage{morewrites}
\usepackage{array}
\usepackage{booktabs}
\usepackage{fvextra}
\usepackage{graphicx}
\usepackage[pdf]{graphviz}
\PassOptionsToPackage{hyphens}{url}
\usepackage[colorlinks=true]{hyperref}
\usepackage{listings}
\usepackage{mdframed}
\usepackage{todonotes}
\usepackage{xcolor}
\usepackage{paralist}

\definecolor{codegreen}{rgb}{0,0.6,0}
\definecolor{codegray}{rgb}{0.5,0.5,0.5}
\definecolor{codepurple}{rgb}{0.58,0,0.82}
\definecolor{backcolour}{rgb}{0.95,0.95,0.92}

\lstdefinelanguage{antlr}{
    keywords = {ACTIVITY,act_fragment},
    keywordstyle=\color{red},
    stringstyle=\color{blue},
    morestring=[b]',
    numbersep=5pt,
}

\mdfdefinestyle{userstorystyle}{
    leftmargin=0cm,%
    rightmargin=0cm,%
    backgroundcolor=black!10,%
    middlelinecolor=black,
    splittopskip=1.5\topsep
}
\newenvironment{userstory}%
    {\begin{mdframed}[style=userstorystyle]}%
    {\end{mdframed}}

\usepackage{cite}
\usepackage{amsmath,amssymb,amsfonts}
\usepackage{algorithmic}
\usepackage{graphicx}
\usepackage{textcomp}
\usepackage{xcolor}
\def\BibTeX{{\rm B\kern-.05em{\sc i\kern-.025em b}\kern-.08em
    T\kern-.1667em\lower.7ex\hbox{E}\kern-.125emX}}

\makeatletter
\newcommand{\linebreakand}{%
  \end{@IEEEauthorhalign}
  \hfill\mbox{}\par
  \mbox{}\hfill\begin{@IEEEauthorhalign}
}
\makeatother

\usepackage{xpatch}
\makeatletter
\newcommand*{\addFileDependency}[1]{
  \typeout{(#1)}
  \@addtofilelist{#1}
  \IfFileExists{#1}{}{\typeout{No file #1.}}
}
\makeatother
\xpretocmd{\digraph}{\addFileDependency{#2.dot}}{}{}

\newcommand{\aname}[1]{\texttt{\color{red}#1}}

\begin{document}

\title{A Framework for Processing Textual Descriptions \\ of Business Processes using a Constrained Language -- Technical Report}

\titlerunning{Processing Textual Descriptions of Business Processes with Constr. Lang.}
\authorrunning{Burattin A. et al.}

\author{Andrea Burattin\inst{1} \and
Antonio Grama\inst{1} \and
Ana-Maria Sima\inst{1} \and
Andrey Rivkin\inst{1} \and
Barbara Weber\inst{2}}

\institute{Technical University of Denmark, DTU Compute, Denmark \\
\email{\{andbur,ariv\}@dtu.dk} \\ \and
University of St.Gallen, St.Gallen, Switzerland\\
\email{barbara.weber@unisg.ch} 
}
\maketitle         

\begin{abstract}    
This report explores how (potentially constrained) natural language can be used to enable nonexperts to develop process models by simply describing scenarios in plain text.
To this end, a framework, called BeePath, is proposed. It allows users to write process descriptions in a constrained pattern-based language, which can then be translated into formal models such as Petri nets and DECLARE. The framework also leverages large language models (LLMs) to help convert unstructured descriptions into this constrained language. 
\end{abstract}

\section{Introduction}

Business processes are often described using natural language. However, natural language introduces ambiguities, inconsistencies, and interpretation gaps that hinder automation and formal verification. Existing process modeling approaches attempt to address these challenges through formal modeling languages, structured annotation techniques, or automated ambiguity resolution. However, these methods either require expertise in formal methods or risk distorting domain experts' intent through rigid rules.
Inspired by structured natural language paradigms such as FRETISH~\cite{GiannakopoulouP21,KatisMGPS22} (a language adopted in NASA’s Formal Requirement Elicitation Tool) and principles from Behavior-Driven Development (BDD), we introduce BeePath, a framework that allows domain experts to describe business processes in a structured, yet natural, way. 

\begin{figure}
    \centering
    \includegraphics[width=\linewidth]{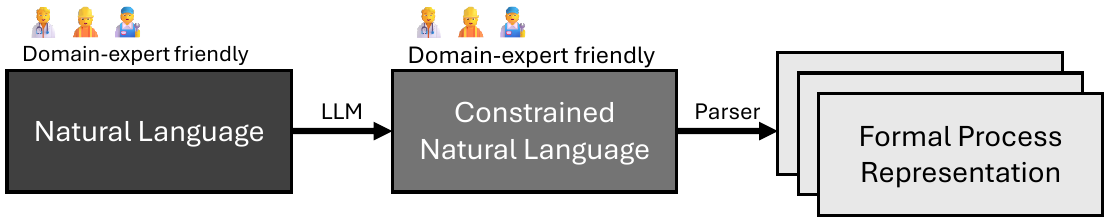}
    \caption{The general pipeline of the BeePath framework.}
    \vspace{-1em}
    \label{fig:idea}
\end{figure}

BeePath follows a two-step approach depicted in Fig.~\ref{fig:idea}:
\begin{itemize}
    \item \textbf{Guided Natural Language Structuring:} Users provide an initial free-text description of a process. This is then transformed into a constrained natural language format, ensuring key constraints and flow dependencies are made explicit, similar to how BDD ensures software behaviors are structured.
    \item \textbf{Model Generation (with Guarantees):} The structured representation is parsed into multiple formal process models, including Petri nets and DECLARE, enabling formal reasoning and automation while maintaining alignment with the original textual intent.
\end{itemize}

Large Language Models (LLMs) are used to assist in the guided natural language structuring, ensuring that process descriptions remain understandable to end users. This allows for human-in-the-loop validation, enabling domain experts to correct any misinterpretations by the LLM. Unlike fully automated approaches, BeePath maintains expert control over the formalization process, ensuring that the structured language output follows a deterministic mapping to process models. This algorithmic mapping prevents hallucinations and guarantees consistency in model generation. A similar approach is seen, for example, in Celonis' LLM for PQL queries\footnote{See \url{https://www.celonis.com/blog/celonis-brings-process-intelligence-to-every-business-user-with-llm-for-pql-generation/}.}. This, however, requires the end-user to know the underlying formal language in order to validate the generated queries.

\section{Constrained Natural Language  for Process Specification}
\label{sec:cnl}

In this section we are going to discuss the grammar of our process description language. We show how this grammar can be used to describe processes and how the obtained descriptions can be automatically translated into Petri nets~\cite{Murata1989} and DECLARE~\cite{DiCiccio2022DeclarativeMonitoring}. 
While the translation into BPMN is also supported in our implementation, it is not covered in this paper due to the page limit and similarity with the Petri net translation.

\subsection{A Language for Describing Processes}
\label{sec:dsl}

\begin{figure}[t!]
    \centering
    \begin{userstory}
        The process starts when the female patient is examined by an outpatient physician, who decides whether she is healthy or needs to undertake an additional examination. In the former case, the physician fills out the examination form and the patient can leave.
        In the latter case, an examination and follow-up treatment order is placed by the physician, who additionally fills out a request form.
        Furthermore, the outpatient physician informs the patient about potential risks.
        If the patient signs an informed consent and agrees to continue with the procedure, a delegate of the physician arranges an appointment of the patient with one of the wards and updates the HIS selecting the first available slot.
        If the patient denies consent the process ends.
        Before the appointment, the required examination and sampling is prepared by a nurse of the ward based on the information provided by the outpatient section.
        Then, a ward physician takes the sample requested.
        He further sends it to the lab indicated in the request form and conducts the follow-up treatment of the patient. After receiving the sample, a physician of the lab validates its state and decides whether the sample can be used for analysis or whether it is contaminated and a new sample is required. 
        After the analysis is performed by a medical technical assistant of the lab, a lab physician validates the results.
        Finally, a physician from the outpatient department makes the diagnosis and prescribes the therapy for the patient.
        
    \end{userstory}
    \caption{Textual description of the ``Hospital process'' from~\cite{Sanchez-Ferreres2019FormalProcesses}.}
    \label{fig:hospital-description}
\end{figure}

The designed process description language, called BeePath, serves as the foundation of the solution, enabling the users to describe processes using a constrained natural language which is much closer to unstructured natural language descriptions.

Conceptually, BeePath relies on having a catalog of textual process description \textit{fragments} available. These fragments can be instantiated several times to achieve the expected final model. We decided to leverage the well-known \textit{workflow patterns} initiative~\cite{Russell2006, VanderAalst2003} for our fragments. These are language-independent workflow structures that capture business characteristics in a generalized manner, making them applicable across various domains.\footnote{See \url{http://www.workflowpatterns.com/patterns/control/} for more detail.}
Our primary goal with BeePath is not to be comprehensive of all possible fragments but rather to define a \textit{framework} that can be easily extended to include more behavior.
For this reason, in addition to the ``Basic Control Flow Patterns''~\cite{VanderAalst2003}, we introduced some additional fragments. To identify these fragments, we collected process descriptions from papers explicitly tackling the problem of modeling from natural language, and we ended up with three descriptions (two from~\cite{friedrich2010}, and one from~\cite{Sanchez-Ferreres2019FormalProcesses} -- cf. Fig.~\ref{fig:hospital-description}). Similarly to pattern elicitation processes~\cite{Lanz2014TimeSystems}, we started with only the Basic Control Flow Patterns, and we carefully inspected the process descriptions to find empirical evidence of new fragments. Given the low number of processes analyzed, we required each candidate fragment to appear once in order to be included on the list. After each new fragment was identified, the process descriptions were analyzed again until no new fragments could be recognized (i.e., saturation).

Please note that the extra fragments identified are not meant to be new \textit{patterns} (i.e., ``reusable solution to a commonly occurring problem''~\cite{Alexander1977ALanguage}). Instead, they are meant to show BeePath's capability to express concepts beyond the Basic Control Flow Patterns.

\begin{figure}[h!]
    \centering
    \includegraphics[width=\linewidth]{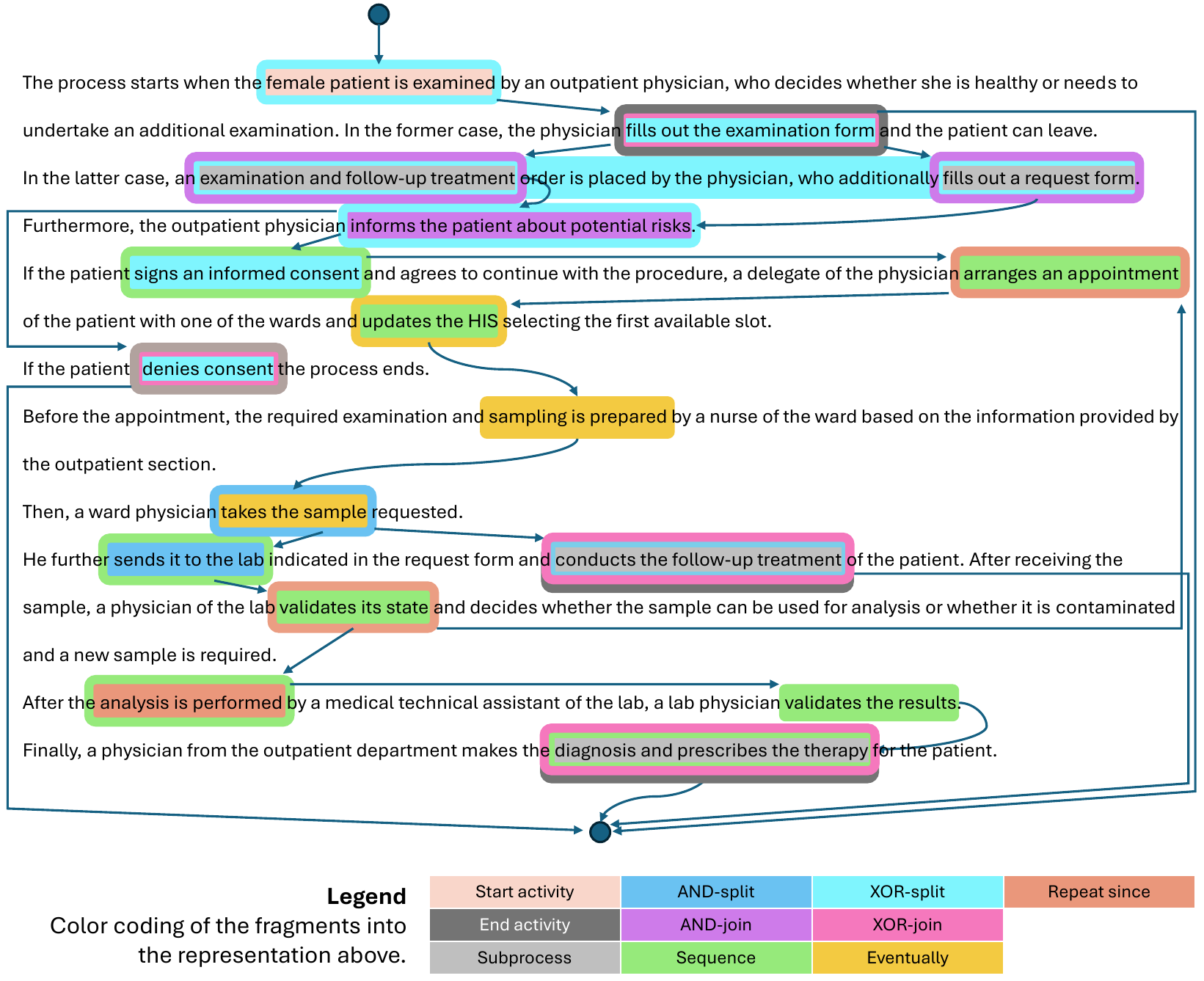}
    \caption{Graphical representation of the Hospital treatment description with the constraints highlighted. Full resolution at \url{https://github.com/dtu-pa/beepath/tree/main/paper-material/hospital-description.pdf}.}
    \label{fig:hospital-annotation}
\end{figure}

Fig.~\ref{fig:hospital-annotation} illustrates the identified fragments within the textual description from Fig.~\ref{fig:hospital-description}. 
In the figure, different text segments are highlighted in colors corresponding to the fragments they represent, as indicated in the legend. 
Additionally, arrows are used to connect fragments and activities visually, enhancing the reader’s understanding. 

\begin{figure*}[th!]
\begin{lstlisting}[language=antlr,basicstyle=\ttfamily\scriptsize]
description: LEADINGTEXT initialStatement fragment+ closingStatement;

LEADINGTEXT: 'The following textual description follows the closed-world assumption, meaning that only...';
initialStatement: 'Initially''start' ACTIVITY;
closingStatement: 'After' ( act_fragment 'ends'
                | act_fragment 'ends' ('and' act_fragment 'ends')+
                | 'either' act_fragment 'ends' ('or' act_fragment 'ends')+
                ) ',''the''process''finishes';
fragment: sequence | parallelSplit | synchronization | exclusiveChoice | simpleMerge | repeatSince | eventually | andSplitInXorSplit | xorSplitInAndSplit | andJoinInXorJoin | xorJoinInAndJoin | subprocess;

sequence          : 'After' ACTIVITY 'ends'',''immediately''start' ACTIVITY;
parallelSplit     : 'After' ACTIVITY 'ends'','
                    'immediately''start' ACTIVITY ('and''start' ACTIVITY)+;
synchronization   : 'After' ACTIVITY 'ends' ('and' ACTIVITY 'ends')+ ','
                    'immediately''start' ACTIVITY;
exclusiveChoice   : 'After' ACTIVITY 'ends'','
                    'immediately''either''start' ACTIVITY 
                    ('or''start' ACTIVITY)+;
simpleMerge       : 'After''either' ACTIVITY 'ends'('or' ACTIVITY 'ends')+','
                    'immediately''start' ACTIVITY;
repeatSince       : 'After' ACTIVITY 'ends'','
                    'immediately''repeat''since' ACTIVITY
                    ('or''start' ACTIVITY)+;
eventually        : 'After' ACTIVITY 'ends'',''eventually''start' ACTIVITY;
andSplitInXorSplit: 'After' ACTIVITY 'ends'','
                    'immediately''either''start' act_fragment
                    ('or''start' act_fragment)+;
xorSplitInAndSplit: 'After' ACTIVITY 'ends'','
                    'immediately''start' act_fragment
                    ('and''start' act_fragment)+;
andJoinInXorJoin  : 'After''either' act_fragment 'ends'
                    ('or' act_fragment 'ends')+ ','
                    'immediately''start' ACTIVITY;
xorJoinInAndJoin  : 'After'act_fragment 'ends'('and' act_fragment 'ends')+','
                    'immediately''start' ACTIVITY;

subprocess: andSubprocess | orSubprocess;
andSubprocess: SUBPROCESS_ID ':' ACTIVITY ('and' ACTIVITY)+;
orSubprocess : SUBPROCESS_ID ':' ACTIVITY ('or' ACTIVITY)+;
act_fragment: ACTIVITY | SUBPROCESS_ID;
ACTIVITY: '"' WORD (' ' WORD)* '"';
SUBPROCESS_ID: '(' WORD+ ')' ;
WORD: ([a-z] | [A-Z] | [0-9] | '_')+;
SPACE: (' ' | '\t' | '.') -> skip;
NEWLINE: ('\r'? '\n' | '\r') -> skip;
\end{lstlisting}
    \caption{The EBNF grammar of the BeePath language presented using ANTLR4 syntax\label{fig:ebnf}}
\end{figure*}

BeePath has been converted into a grammar that includes all the identified process fragments. This grammar is described using extended Backus–Naur form (EBNF), and is available in Fig~\ref{fig:ebnf}.
The main production of the grammar is the following:
\begin{lstlisting}[language=antlr]
description: LEADINGTEXT
             initialStatement
             fragment+
             closingStatement;
\end{lstlisting}
According to the above rule, a process description must start with a  \lstinline{LEADINGTEXT}, followed by an \lstinline{initialStatement}, at least one \lstinline{fragment}, and a \lstinline{closingStatement}.
\lstinline{LEADINGTEXT} plays the role of a ``disclaimer''-like statement and aims at making biases applied in the interpretation of BeePath descriptions explicit, thus making the user aware of how such descriptions will be interpreted when compiled into a target formal model using parsers supported by the framework (the parsers are described in Section~\ref{sec:parsing}). 
Currently, the following leading text is provided: {\color{blue}\footnotesize \texttt{The following textual description follows the closed-world assumption, meaning that only the activities specified can be executed in the specified order. Any possible activity and execution that is not specified is considered\\ impossible}}.

The \lstinline{initialStatement} defines the initial activity. The possibility of allowing only one activity has been decided based on observations in the sample textual descriptions.
This restriction can be easily lifted to the set of initial activities by modifying the \lstinline{initialStatement} production. 
The \lstinline{closingStatement} production indicates the activities leading to a conclusion of the process. There is no restriction on the number of these, and the ``process finishes'' phrase is used due to being a common phrase identified in the analysed three textual descriptions of business processes.

The list of \lstinline{fragment}s is defined over these non-terminals:
\begin{lstlisting}[language=antlr]
fragment: sequence | parallelSplit | synchronization
        | exclusiveChoice | simpleMerge
        | repeatSince | eventually
        | andSplitInXorSplit | xorSplitInAndSplit
        | andJoinInXorJoin | xorJoinInAndJoin
        | subprocess;
\end{lstlisting}

\begin{table*}
    \caption{Instances of identified fragments and corresponding strings generated in the BeePath language. Fragment arguments appear in {\color{red}red}.}
    \label{tab:patterns}
    \centering
    \resizebox{.99\textwidth}{!}{
        \begin{tabular}{ll}
            \toprule
            \textbf{Fragment instance} & \multicolumn{1}{l}{\textbf{Example of a BeePath language string}} \\
            \midrule
            
            \textsf{Sequence\,(\aname{A}, \aname{B})} & \lstinline{After "A" ends, immediately start "B".} \\
            \textsf{Parallel Split\,(\aname{A}, \aname{B}, \aname{C})} & \lstinline{After "A" ends, immediately start "B" and start "C".} \\
            \textsf{Synchronization\,(\aname{A}, \aname{B}, \aname{C})}  & \lstinline{After "A" ends and "B" ends, immediately start "C".} \\
            \textsf{Exclusive Choice\,(\aname{A}, \aname{B}, \aname{C})} & \lstinline{After "A" ends, immediately either start "B" or start "C".} \\
            \textsf{Simple Merge\,(\aname{A}, \aname{B}, \aname{C})} & \lstinline{After either "A" ends or "B" ends, immediately start "C".} \\[.5em]
            
            \textsf{Repeat Since\,(\aname{A}, \aname{B}, \aname{C})} & \lstinline{After "A" ends, immediately repeat since "B" or start "C".} \\
            \textsf{Eventually\,(\aname{A}, \aname{B})} & \lstinline{After "A" ends, eventually start "B".} \\
            
            \textsf{AND-split in XOR-split\,(\aname{A}, \aname{B\_and\_C}, \aname{D})} & \lstinline{After "A" ends, immediately either start (B\_and\_C) or start "D".} \\
            \textsf{AND-join in XOR-join\,(\aname{A\_and\_B}, \aname{C}, \aname{E})} & \lstinline{After either (A\_and\_B) ends or "C" ends, immediately start "E".} \\
            \textsf{XOR-split in AND-split\,(\aname{A}, \aname{B\_or\_C}, \aname{D})} & \lstinline{After "A" ends, immediately start (B\_or\_C) and start "D".} \\
            \textsf{XOR-join in AND-join\,(\aname{A\_or\_B}, \aname{C}, \aname{E})} & \lstinline{After (A\_or\_B) ends and "C" ends, immediately start "E".} \\[.5em]
            
            \textsf{OR-subprocess\,(\aname{A\_or\_B}, \aname{A}, \aname{B})} & \lstinline{(A\_or\_B): "A" or "B"} \\
            \textsf{AND-subprocess\,(\aname{A\_and\_B}, \aname{A}, \aname{B})} & \lstinline{(A\_and\_B): "A" and "B"} \\
            \bottomrule
        \end{tabular}
    }
\end{table*}

Examples of strings derived from the list of fragment productions can be seen in Table~\ref{tab:patterns}, where the first column states the fragment name  and the second provides a corresponding BeePath string.
As the table caption also suggests, it is easy to see fragments having formal parameters referring to activities and/or sub-process identifiers. It is easy to see from the grammar in Figure~\ref{fig:ebnf} that almost all the fragments can have a variable number of parameters.
On top of that, all the fragments refer to the activities' start and end events.
Below, we discuss three selected fragment productions from the complete grammar in detail.

The \lstinline{sequence} production corresponds to the sequence pattern~\cite{VanderAalst2003}, which specifies two consecutive activity executions, where the following activity can start only after the preceding one has ended. In the BeePath grammar:
\begin{lstlisting}[language=antlr]
sequence: 'After' ACTIVITY 'ends'','
          'immediately''start' ACTIVITY;
\end{lstlisting}
It is also important to notice that the \lstinline{sequence} production is defined only for pairs of activities. This, however, is not a limitation (and purely a design choice) as more than two activities can be aptly organized in one sequence by writing a series of \lstinline{sequence} strings.

The \lstinline{parallelSplit} production allows to define generate descriptions describing scenarios in which termination of one activity follows by an immediate start of two or more activities that are supposed to be executed in parallel:
\begin{lstlisting}[language=antlr]
parallelSplit: 'After' ACTIVITY 'ends'','
               'immediately''start' ACTIVITY
               ('and''start' ACTIVITY)+;
\end{lstlisting}
Notice that the pattern modelled with this production corresponds to AND-split constructs adopted in BPM~\cite{VanderAalst2003,Dumas2013a}.

To describe the synchronisation of finished activities that were running in parallel,
the  \lstinline{synchronization} production is introduced:
\begin{lstlisting}[language=antlr]
synchronization: 'After' ACTIVITY 'ends'
                 ('and' ACTIVITY 'ends')+ ','
                 'immediately''start' ACTIVITY;
\end{lstlisting}
Similarly to the previous case, this production models AND-join constructs~\cite{VanderAalst2003,Dumas2013a}.

Analogously, productions \lstinline{exclusiveChoice} and \lstinline{simpleMerge}
respectively account for XOR-splits and -joins~\cite{VanderAalst2003,Dumas2013a}. 

As already mentioned, the above productions generate descriptions corresponding to the \textit{sequence}, \textit{parallel split}, \textit{synchronization}, \textit{exclusive choice}, and \textit{simple merge}, i.e., the Basic Control Flow Patterns~\cite{VanderAalst2003}. As can be seen from the listed productions and the BeePath grammar, there is no limit on the number of ``branches'' of splits and merges (even though the examples in Table~\ref{tab:patterns} only report two).

Other two fragments identified during the analytic phase are useful to capture repetitions of activities and non-immediate succession relations between them. 
To this end, \textit{Repeat since} and \textit{eventually} are introduced.

Next to these, from the textual descriptions, we identified that parallel splits and exclusive choices can be expressed in a compact way in a single statement. For this reason, we decided to instrument our BeePath grammar with these language primitives that allow to both split and join the behavior (\textit{AND-split in XOR-split} and \textit{AND-join in XOR-join}). For symmetry purposes and leveraging the extensibility of BeePath we also decided to add two extra fragments (\textit{XOR-split in AND-split} and \textit{XOR-join in AND-join}). 
Finally, to simplify the textual representation and avoid duplication of activities, we introduce \textit{subprocesses} that can express exclusive or concurrent behavior of at least two activities. These subprocesses, which are defined separately, can be used inside the other fragments.

Considering the process described in Fig.~\ref{fig:hospital-description}, a possible translation using the BeePath language is presented in Fig.~\ref{lst:hospital-beepath}.

\begin{figure*}
\begin{lstlisting}[basicstyle=\ttfamily\scriptsize]
The following textual description follows the closed-world assumption, meaning that only the activities...
Initially start "examine female patient".
(s1): "order examination with follow up treatment" and "fill out request form".
(s2): "conduct follow up treatment" and "diagnose with prescribe therapy".
After "examine female patient" ends, immediately either start "fill out examination form" or start (s1).
After "order examination with follow up treatment" ends and "fill out request form" ends, immediately start "inform patient about risks".
After "inform patient about risks" ends, immediately either start "sign consent" or start "deny consent".
After "sign consent" ends, immediately start "arrange appointment".
After "arrange appointment" ends, immediately start "update HIS selecting".
After "update HIS selecting" ends, eventually start "prepare sampling".
After "prepare sampling" ends, eventually start "take sample".
After "take sample" ends, immediately start "send sample to lab" and start "conduct follow up treatment".
After "send sample to lab" ends, immediately start "validate sample".
After "validate sample" ends, immediately repeat since "arrange appointment" or start "perform analysis".
After "perform analysis" ends, immediately start "validate results".
After "validate results" ends, immediately start "diagnose with prescribe therapy".
After either (s2) ends or "deny consent" ends or "fill out examination form" ends, the process finishes.
\end{lstlisting}
    \vspace{-1em}
    \caption{Conversion in BeePath language of the Hospital treatment example.\label{lst:hospital-beepath}}
\end{figure*}

Let us now provide a few observations on the BeePath language. 
First of all, the BeePath language cannot automatically resolve (semantic) ambiguities present in source texts but can offer a structured approach to force users to do so.
Second, the language also offers ways to deal with incomplete (source) specifications, including cases of missing or not clearly stated connections between activities.
For instance, in Fig.~\ref{fig:hospital-annotation}, if a ``sample analysis'' was deemed invalid, the original text only mentioned that ``a new sample was required'' without specifying which activity should follow (e.g., ``take sample'' or ``arrange appointment''). 
However, when converting the description to BeePath, the user is required to explicitly resolve such issues by specifying which activity should be repeated. Without this, the resulting model would be incomplete and disconnected (the latter can be generated by the algorithms discussed in Sect.~\ref{sec:parsing}).

The version of the BeePath language reported in this paper exhibits the following limitations:
\begin{inparaenum}[(1)]

\item Imperative bias. The current BeePath grammar has an imperative bias, which is ``reinforced'' by the closed-world assumption adopted in its design. This means that, when describing a process in BeePath, all possible execution paths must be explicitly defined (as per the current ``Leading Text''). This contrasts with a declarative approach, where only constraints must be specified, and makes the language less suitable for highly flexible processes with numerous execution paths (as descriptions can become lengthy, complex, overly detailed, and challenging to maintain, increasing the risk of errors).
\item The lack of nested subprocesses. The current BeePath grammar does not allow to specify nested subprocesses (e.g., an AND subprocess within an OR subprocess). This restriction was a deliberate choice to avoid adding unnecessary complexity to the grammar, underlying patterns, and their translation into formal process models.
\item The lack of (process) resources and roles. The current version of the BeePath grammar does not provide a way to specify details about actors, timestamps, conditions, or additional annotations. Although some of this information appears in the sample textual descriptions. 
\end{inparaenum}

\subsection{Parsing BeePath}
\label{sec:parsing}
Once the textual description is available, the parsing algorithm is executed. The parsing algorithm has the goal of transforming sentences written in the BeePath language into a target (formal) process model expressed in one of the supported languages. In a nutshell, the parsing algorithm goes through the following steps:
\begin{compactenum}[\bfseries (1)]
    \item Iterate through all input sentences in the order in which they are written and generate the corresponding abstract syntax tree (AST). 
    \item Use the AST to identify for each sentence which fragment type it represents and its arguments (activity names and/or sub-process identifiers). Store the obtained information in a ``global storage''.\footnote{This is needed to keep the information about encountered activities in order to maintain the referential integrity between parsed sentences (that is, sentences that ``re-use'' the same activities will be parsed into formal models with the same activity labels)}
    \item For each identified fragment, invoke a corresponding translation sub-routine yielding a fragment of the target formal model. Such sub-routines are discussed below.
    \item Produce the global model by aptly joining the outputs for all identified fragments.\footnote{The join operation is specific to each target formalism.} 
\end{compactenum}

We now discuss which formalisms are currently supported in the parsing algorithm.

\subsubsection{Supported Modeling Paradigms}
In the current implementation of the BeePath framework, the parsing is targeting two languages: Petri nets~\cite{Peterson1977,Reisig2013UnderstandingNets} and DECLARE~\cite{Pesic2006,DiCiccio2022DeclarativeMonitoring}. We opted for these two languages as representative of two process modeling paradigms: imperative and declarative.

\begin{itemize}
    \item \textbf{Imperative Modelling (Petri nets).}
    In imperative modeling, it is required that the entire procedure for executing a process instance is explicitly defined. Like that all possible execution paths are specified in the model before the process runs, which often leads to overspecified process models~\cite{Pichler2012}.
    In this work, we use Petri nets as the imperative formalism.\footnote{To be precise, we rely on standard P/T nets ~\cite{Murata1989} without arc weights.} 
    We refer to~\cite{Murata1989} for detailed definitions of the Petri net syntax and semantics. 
    In our translation, each activity \textsf{A} is mapped into two Petri net transitions, one indicating the beginning of the activity (labeled with \textsf{A\_start}) and the other modeling its end (labeled with \textsf{A\_end}).
    In addition to labeled transitions, some of the patterns discussed in this work also utilize \textit{silent transitions}, i.e., ``system'' transitions not associated with any label and that help modeling the target behavior.
    Graphically, such transitions appear as black rectangles.
    \item \textbf{Declarative Modelling (DECLARE).}
In declarative modeling, one  defines a set of process activities in focus alongside a set of constraints (that use such activities), avoiding an explicit specification of how the process shuld be executed. In this paper, we use DECLARE as the declarative process modeling language~\cite{DiCiccio2022DeclarativeMonitoring}. In DECLARE models, any process execution is allowed as long as it does not violate the specified constraints. Compared to imperative modeling, declarative modeling is more flexible and imposes fewer restrictions on process specifications, making it particularly suited for loosely structured and highly adaptable processes. 
We refer to~\cite{DiCiccio2022DeclarativeMonitoring} for the complete formalisation of DECLARE, including the compelte set of supported templates and the language semantics. 
Since DECLARE is a template-based language in which each instantiated template (or their collection) can be represented as finite-state automata ~\cite{GiacomoV13,DiCiccio2022DeclarativeMonitoring}, in this paper we use both template- and automata-based representations for the DECLARE translation of the patterns supported by the BeePath language. 
It is also important to note that our translation into DECLARE focuses solely on when activities start, which is in line  with the activation concept adopted in the DECLARE language.~\cite{DiCiccio2022DeclarativeMonitoring}.
\end{itemize}

\subsubsection{Mapping BeePath Fragments into Formal Models}

We now describe how the fragments identified in Section~\ref{sec:dsl} are translated into Petri nets and DECLARE specifications. 
Notice that focusing solely on such mappings is sufficient, as the parsing algorithm described above inherently relies on identifying fragments (along with other relevant information) in a given set of strings and subsequently translating the fragments into formal models, which are then combined into the final resulting model.

For better readability, we will show how instances listed in Table~\ref{tab:patterns} are translated. Notice that the supplied implementation (\url{https://github.com/dtu-pa/beepath/}) translates the most general fragments, i.e., those coming with an arbitrary number of activities.

\smallskip
\noindent\textsf{Sequence\,(\aname{A}, \aname{B})}. This fragment is translated into the following Petri net $\mathcal{N}$:
\begin{center}
    \digraph[scale=0.6]{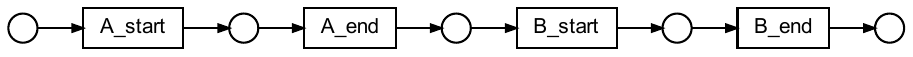}{ranksep=".3"; fontsize="10"; remincross=true; margin="0.0,0.0"; fontname="Arial"; rankdir="LR";
        edge [arrowsize="0.5"];
	node [height=".2",width=".2",fontname="Arial",fontsize="10"];
	"1" [shape="circle",label=""];
	"2" [shape="circle",label=""];
	"3" [shape="circle",label=""];
	"4" [shape="circle",label=""];
	"5" [shape="circle",label=""];
	"A_start" [shape="box"];
	"1" -> "A_start";
	"A_start" -> "2";
	"A_end" [shape="box"];
	"2" -> "A_end";
	"A_end" -> "3";
	"B_start" [shape="box"];
	"3" -> "B_start";
	"B_start" -> "4";
	"B_end" [shape="box"];
	"4" -> "B_end";
	"B_end" -> "5";}
\end{center}
$\mathcal{N}$ ensures that the \textit{end} transition of one activity (e.g., \textsf{A\_end}) is directly connected to the \textit{start} transition of the next activity (e.g., \textsf{B\_start}), with no other transitions occurring between them. This connection is established through an intermediary place. 
The DECLARE specification $\mathcal{D}$ for this fragment and its corresponding automaton are as follows:
    
        \begin{align*}
            \mathcal{D}=\{\,\textsf{Chain Succession(A, B)}\,\}
        \end{align*}
\begin{center}

    \digraph[scale=0.6]{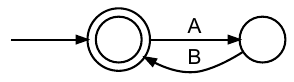}{margin="0.0,0.0"; rankdir="LR";
        edge [arrowsize="0.5"];
        start [label="",style = "invisible",width=0, height=0];
	node [height=".3",width=".3",shape="circle",label=""];
	2 [peripheries=2];
	1;
	start->2;
	2->1 [label="A",fontname="Arial",fontsize="10"];
	1->2 [label="B",fontname="Arial",fontsize="10"];}%
\end{center}

\noindent
\textsf{Parallel Split\,(\aname{A}, \aname{B}, \aname{C})} and \textsf{Synchronization\,(\aname{A}, \aname{B}, \aname{C})}.
These two complementary fragments are translated into two Petri nets $\mathcal{N}_1$ and $\mathcal{N}_2$:
\begin{center}
    \digraph[scale=0.6]{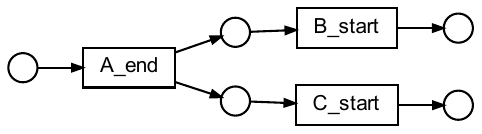}{ranksep=".3"; fontsize="10"; remincross=true; margin="0.0,0.0"; fontname="Arial"; rankdir="LR";
        edge [arrowsize="0.5"];
	node [height=".2",width=".2",fontname="Arial",fontsize="10"];
	"1" [shape="circle",label=""];
	"2" [shape="circle",label=""];
	"3" [shape="circle",label=""];
	"4" [shape="circle",label=""];
	"5" [shape="circle",label=""];
	"A_end" [shape="box",label="A_end"];
	"1" -> "A_end";
	"A_end" -> "2";
	"A_end" -> "3";
	"B_start" [shape="box",label="B_start"];
	"2" -> "B_start";
	"B_start" -> "4";
	"C_start" [shape="box",label="C_start"];
	"3" -> "C_start";
	"C_start" -> "5";}%
\digraph[scale=0.6]{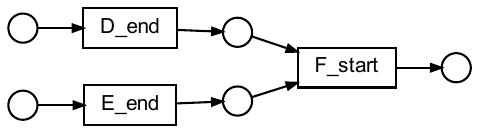}{ranksep=".3"; fontsize="10"; remincross=true; margin="0.0,0.0"; fontname="Arial"; rankdir="LR";
        edge [arrowsize="0.5"];
	node [height=".2",width=".2",fontname="Arial",fontsize="10"];
	"1" [shape="circle",label=""];
	"2" [shape="circle",label=""];
	"3" [shape="circle",label=""];
	"4" [shape="circle",label=""];
	"5" [shape="circle",label=""];
	"D_end" [shape="box"];
        "E_end" [shape="box"];
        "F_start" [shape="box"];
	"1" -> "D_end" -> "2" -> "F_start";
        "3" -> "E_end" -> "4" -> "F_start";
        "F_start" -> "5";}
\end{center}   
$\mathcal{N}_1$ ensures that multiple activities or subprocesses can start simultaneously by connecting their \textit{start} transitions (e.g., \textsf{B\_start} and \textsf{C\_start}) to the \textit{end} transition of the preceding activity (e.g., \textsf{A\_end}), with no other transitions occurring between them. $\mathcal{N}_2$ ensure that the \textit{end} transitions of multiple activities (e.g., \textsf{D\_end} and \textsf{E\_end}) or subprocesses are all connected to the \textit{start} transition of the next activity (e.g., \textsf{F\_start}), with no other transitions occurring between them.
The DECLARE specification $\mathcal{D}_1$ for \textsf{Parallel Split} is the following:
        \begin{align*}
            \mathcal{D}_1=\{\,&\textsf{Alternate Succession(A, B)}; \textsf{Alternate Succession(A, C)};\\ 
                          &\textsf{Exactly One(A)}\,\}
        \end{align*}
\begin{center}

    \digraph[scale=0.6]{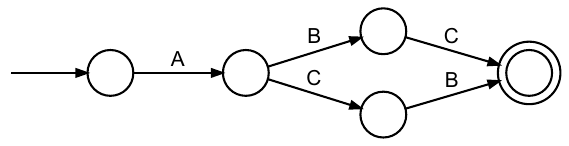}{margin="0.0,0.0"; rankdir="LR";
        edge [arrowsize="0.5"];
        start [label="",style = "invisible",width=0, height=0];
	node [height=".3",width=".3",shape="circle",label=""];
        1; 2; 3; 4;
        5 [peripheries=2];
	start->1;
	1->2 [label="A",fontname="Arial",fontsize="10"];
	2->3 [label="B",fontname="Arial",fontsize="10"];
        2->4 [label="C",fontname="Arial",fontsize="10"];
        3->5 [label="C",fontname="Arial",fontsize="10"];
        4->5 [label="B",fontname="Arial",fontsize="10"];}%
\end{center} 
The DECLARE representation  $\mathcal{D}_2$ for \textsf{Synchronisation} is the following:
        \begin{align*}
            \mathcal{D}_2=\{\,&\textsf{Alternate Succession(D, F)}; \textsf{Alternate Succession(E, F)};\\ 
                          &\textsf{Exactly One(F)}\,\}
        \end{align*}
\begin{center}

    \digraph[scale=0.6]{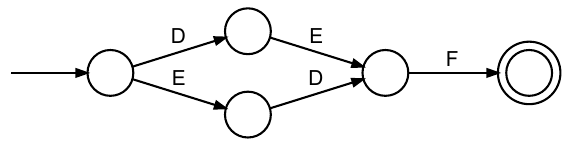}{margin="0.0,0.0"; rankdir="LR";
        edge [arrowsize="0.5"];
        start [label="",style = "invisible",width=0, height=0];
	node [height=".3",width=".3",shape="circle",label=""];
        5; 2; 3; 4;
        1 [peripheries=2];
	start->2;
	2->3 [label="D",fontname="Arial",fontsize="10"];
        2->4 [label="E",fontname="Arial",fontsize="10"];
        3->5 [label="E",fontname="Arial",fontsize="10"];
        4->5 [label="D",fontname="Arial",fontsize="10"];
        5->1 [label="F",fontname="Arial",fontsize="10"];}%
\end{center}

\noindent
\textsf{Exclusive Choice\,(\aname{A}, \aname{B}, \aname{C})} and \textsf{Simple Merge\,(\aname{D}, \aname{E}, \aname{F})}.
These two complementary fragments are translated into two Petri nets $\mathcal{N}_1$ and $\mathcal{N}_2$:

\begin{center}
    \digraph[scale=0.6]{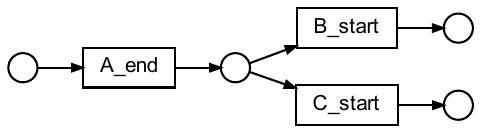}{ranksep=".3"; fontsize="10"; remincross=true; margin="0.0,0.0"; fontname="Arial"; rankdir="LR";
        edge [arrowsize="0.5"];
	node [height=".2",width=".2",fontname="Arial",fontsize="10"];
	"1" [shape="circle",label=""];
	"2" [shape="circle",label=""];
	"4" [shape="circle",label=""];
	"5" [shape="circle",label=""];
	"A_end" [shape="box"];
	"1" -> "A_end";
	"A_end" -> "2";
	"B_start" [shape="box"];
	"2" -> "B_start";
	"B_start" -> "4";
	"C_start" [shape="box"];
	"2" -> "C_start";
	"C_start" -> "5";}%
\digraph[scale=0.55]{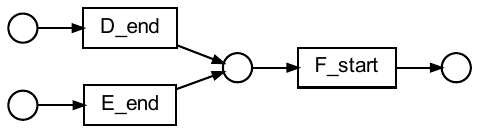}{ranksep=".3"; fontsize="10"; remincross=true; margin="0.0,0.0"; fontname="Arial"; rankdir="LR";
        edge [arrowsize="0.5"];
	node [height=".2",width=".2",fontname="Arial",fontsize="10"];
	"1" [shape="circle",label=""];
	"2" [shape="circle",label=""];
	"3" [shape="circle",label=""];
	"5" [shape="circle",label=""];
	"D_end" [shape="box"];
        "E_end" [shape="box"];
        "F_start" [shape="box"];
	"1" -> "D_end" -> "2" -> "F_start";
        "3" -> "E_end" -> "2";
        "F_start" -> "5";}
\end{center} 
$\mathcal{N}_1$ ensures that when an activity ends, its \textit{end} transition (e.g., \textsf{A\_end})is connected to the \textit{start} transitions (e.g., \textsf{B\_start} and \textsf{C\_start}) of multiple alternative activities or subprocesses, with only one of them being able to execute. No other transitions can occur between them.
$\mathcal{N}_2$ ensures that multiple alternative activities or subprocesses can \textit{end} (e.g., \textsf{D\_end} and \textsf{E\_end}), with only one of them leading to the \textit{start} of the next activity (e.g., \textsf{F\_start}).
The DECLARE representation $\mathcal{D}_1$ for \textsf{Exclusive Choice} requires a complex combination of DECLARE templates, including an ``artificial'' end event that forces $\mathcal{D}_1$ to complete, even though their corresponding automata-based representation is straightforward:
        \begin{align*}
            \mathcal{D}_1=\{\,&
                \textsf{Not CoExistence(B, C)};
                \textsf{Alternate Precedence(B, A)};\\ 
                &\textsf{Alternate Precedence(C, A)};
                \textsf{Exactly One(A)};\\
                &\textsf{Alternate Response(B, end)};
                \textsf{Alternate Response(C, end)};\\
                &\textsf{Not Chain Succession(A, end)};
                \textsf{Not Chain Succession(end, A)}
            \,\}
        \end{align*}
\begin{center}

    \digraph[scale=0.6]{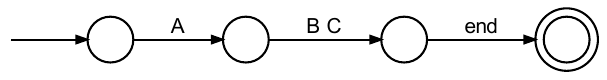}{margin="0.0,0.0"; rankdir="LR";
        edge [arrowsize="0.5"];
        start [label="",style = "invisible",width=0, height=0];
	node [height=".3",width=".3",shape="circle",label=""];
        1; 2; 3;
        4 [peripheries=2];
	start->1;
	1->2 [label="A",fontname="Arial",fontsize="10"];
	2->3 [label="B C",fontname="Arial",fontsize="10"];
        3->4 [label="end",fontname="Arial",fontsize="10"];}%
\end{center}
The DECLARE representation $\mathcal{D}_2$ for \textsf{Simple Merge} requires the dual ``artificial'' initialisation event (marked as \textsf{init}) and is as follows:
        \begin{align*}
            \mathcal{D}_2=\{\,&
                \textsf{Alternate Response(D, F)};
                \textsf{Alternate Response(E, F)};\\
                &\textsf{Not CoExistence(D, E)};
                \textsf{Not Chain Succession(init, F)};\\
                &\textsf{Not Chain Succession(F, init)};
                \textsf{Exactly One(F)};\\
                &\textsf{Init(init)};
                \textsf{Exactly One(init)}
            \,\}
        \end{align*}
\begin{center}

    \digraph[scale=0.6]{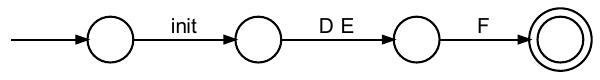}{margin="0.0,0.0"; rankdir="LR";
        edge [arrowsize="0.5"];
        start [label="",style = "invisible",width=0, height=0];
	node [height=".3",width=".3",shape="circle",label=""];
        1; 2; 3;
        4 [peripheries=2];
	start->1;
	1->2 [label="init",fontname="Arial",fontsize="10"];
	2->3 [label="D E",fontname="Arial",fontsize="10"];
        3->4 [label="F",fontname="Arial",fontsize="10"];}%
\end{center}

\noindent
\textsf{Repeat Since(\aname{A}, \aname{B}, \aname{C})}. It translates into the following Petri net $\mathcal{N}$:
\begin{center}
    \digraph[scale=0.6]{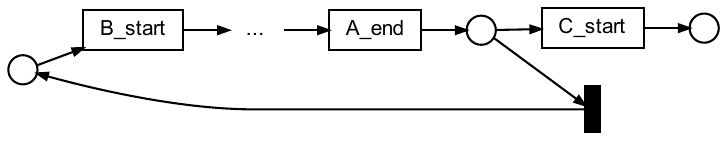}{ranksep=".3"; fontsize="10"; remincross=true; margin="0.0,0.0"; fontname="Arial"; rankdir="LR";
        edge [arrowsize="0.5"];
	node [height=".2",width=".2",fontname="Arial",fontsize="10"];
	"1" [shape="circle",label=""];
	"2" [shape="circle",label=""];
	"3" [shape="circle",label=""];
	"A_end" [shape="box"];
	"C_start" [shape="box"];
	"B_start" [shape="box"];
	"1" -> "B_start";
        "A_end" -> "2" ->"C_start" -> "3";
        dots [label="...",shape="plaintext"];
        silent [shape="box",label="",height=".3",width=".1",style="filled",fillcolor="black"];
        "B_start"->dots->"A_end";
        "2" -> silent -> "1";}
\end{center}
$\mathcal{N}$ ensures that when an activity \textit{ends} (e..g, \textsf{A\_end)}, 
the execution can continue to either the \textit{start} transition of the activity that needs to be repeated (e..g, \textsf{B\_start)} or to the \textit{start} transitions of any new starting activities or subprocess (e..g, \textsf{C\_start)}. No other transitions can occur between them.
We decided to have an empty DECLARE translation for \textsf{Repeat Since} as a declarative approach does not require specifying the possibility of repeating parts of the process (this possibility is in the nature of declarative modelling).

\noindent
\textsf{Eventually(\aname{A}, \aname{B})}. The translation into Petri nets is similar to \textsf{Sequence(\aname{A}, \aname{B})} with the only difference that \textit{end} and \textit{start} transitions can contain other transitions in between. 
The DECLARE representation $\mathcal{D}$ of \textsf{Eventually}, instead, is as follows:
        \begin{align*}
            \mathcal{D}=\{\,\textsf{Succession(A, B)}\,\}
        \end{align*}
\begin{center}

    \digraph[scale=0.6]{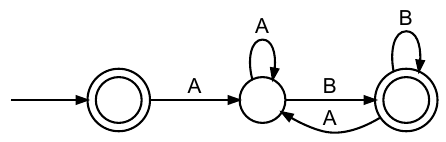}{margin="0.0,0.0"; rankdir="LR";
        edge [arrowsize="0.5"];
        start [label="",style = "invisible",width=0, height=0];
	node [height=".3",width=".3",shape="circle",label=""];
        1 [peripheries=2];
        2;
        3 [peripheries=2];
	start->1;
	1->2 [label="A",fontname="Arial",fontsize="10"];
        2->2 [label="A",fontname="Arial",fontsize="10"];
	2->3 [label="B",fontname="Arial",fontsize="10"];
        3->3 [label="B",fontname="Arial",fontsize="10"];
        3->2 [label="A",fontname="Arial",fontsize="10"];}
\end{center}

\noindent
\textsf{AND-split in XOR-split\,(\aname{A}, \aname{B\_and\_C}, \aname{D})} and 
\textsf{AND-join in XOR-join\,(\aname{E\_and\_F}, \aname{G}, \aname{H})}. 
The translation into corresponding Petri nets $\mathcal{N}_1$ and $\mathcal{N}_2$ is as follows:
\\
\hspace{1em}
\digraph[scale=0.6]{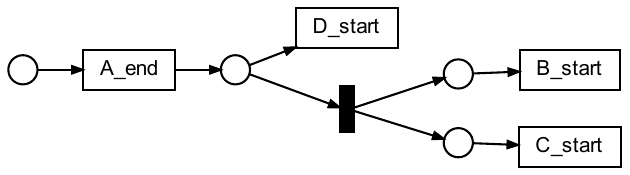}{ranksep=".3"; fontsize="10"; remincross=true; margin="0.0,0.0"; fontname="Arial"; rankdir="LR";
        edge [arrowsize="0.5"];
	node [height=".2",width=".2",fontname="Arial",fontsize="10"];
	"1" [shape="circle",label=""];
	"2" [shape="circle",label=""];
	"3" [shape="circle",label=""];
	"4" [shape="circle",label=""];
	"A_end" [shape="box"];
	"B_start" [shape="box"];
	"C_start" [shape="box"];
	"D_start" [shape="box"];
        silent [shape="box",label="",height=".3",width=".1",style="filled",fillcolor="black"];
	"1" -> "A_end";
	"A_end" -> "2";
	"2" -> silent;
	silent -> "3" -> "B_start";
	silent -> "4" -> "C_start";
	"2" -> "D_start";}
\begin{flushright}
\digraph[scale=0.6]{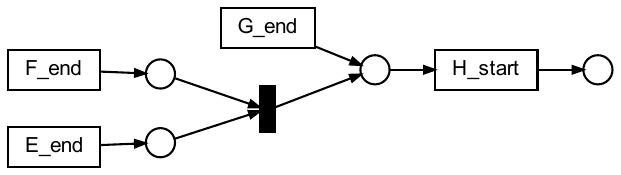}{ranksep=".3"; fontsize="10"; remincross=true; margin="0.0,0.0"; fontname="Arial"; rankdir="LR";
        edge [arrowsize="0.5"];
	node [height=".2",width=".2",fontname="Arial",fontsize="10"];
	"1" [shape="circle",label=""];
	"2" [shape="circle",label=""];
	"3" [shape="circle",label=""];
	"4" [shape="circle",label=""];
	"E_end" [shape="box"];
	"F_end" [shape="box"];
	"G_end" [shape="box"];
	"H_start" [shape="box"];
        silent [shape="box",label="",height=".3",width=".1",style="filled",fillcolor="black"];
	"E_end" -> "1" -> silent;
	"F_end" -> "2" -> silent;
	silent -> "3" -> "H_start" -> "4";
	"G_end" -> "3";}\\
\end{flushright}
The difference between an exclusive choice and an XOR-split (which contains an AND-split inside) lies in the handling of AND subprocesses. In a standard choice, only one of the possible starting activities or subprocesses is chosen after the \textit{ending} activity finishes (e.g., \textsf{A\_end}). In a nested XOR-split, the AND-split is embedded within one or more branches, meaning that, possibly, multiple activities or subprocesses may \textit{start} simultaneously (e.g., \textsf{B\_start} and \textsf{C\_start}). Despite this, the core behavior remains the same.
The dual situation is there for the AND-join inside of an XOR-join.
The DECLARE representation $\mathcal{D}_1$ of \textsf{AND-split in XOR-split} requires a complex combination of templates, including an ``artificial'' \textsf{end} activity that forces $\mathcal{D}_1$ to complete:
        \begin{align*}
            \mathcal{D}_1=\{\,&
                \textsf{Alternate Precedence(B, A)};
                \textsf{Alternate Precedence(C, A)};\\
                &\textsf{Alternate Precedence(D, A)};
                \textsf{CoExistence(B, C)};\\
                &\textsf{Not CoExistence(B, D)};
                \textsf{Alternate Response(B, end)};\\
                &\textsf{Alternate Response(C, end)};
                \textsf{Alternate Response(D, end)};\\
                &\textsf{Exactly One(A)};
                \textsf{Exactly One(end)};\\
                &\textsf{Not Chain Succession(A, end)};
                \textsf{Not Chain Succession(end, A)}
            \,\}
        \end{align*}
\begin{center}

    \digraph[scale=0.6]{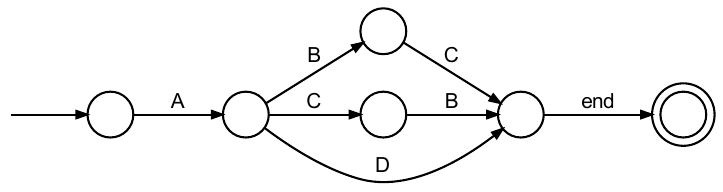}{margin="0.0,0.0"; rankdir="LR";
        edge [arrowsize="0.5"];
        start [label="",style = "invisible",width=0, height=0];
	node [height=".3",width=".3",shape="circle",label=""];
        1; 2; 3; 4; 5;
        6 [peripheries=2];
	start->1;
	1->2 [label="A",fontname="Arial",fontsize="10"];
	2->3 [label="B",fontname="Arial",fontsize="10"];
        3->5 [label="C",fontname="Arial",fontsize="10"];
        2->4 [label="C",fontname="Arial",fontsize="10"];
        4->5 [label="B",fontname="Arial",fontsize="10"];
        2->5 [label="D",fontname="Arial",fontsize="10"];
        5->6 [label="end",fontname="Arial",fontsize="10"];
        }
\end{center}
The DECLARE representation $\mathcal{D}_2$ of \textsf{AND-join in an XOR-join} requires the dual ``artificial'' \textsf{init} activity and is as follows:
        \begin{align*}
            \mathcal{D}_2=\{\,&
                \textsf{Alternate Response(E, H)};
                \textsf{Alternate Response(F, H)};\\
                &\textsf{Alternate Response(G, H)};
                \textsf{CoExistence(E, F)};\\
                &\textsf{Not CoExistence(E, G)};
                \textsf{Exactly One(H)};
                \textsf{Init(init)}; \\
                &\textsf{Not Chain Succession(init, e)};
                \textsf{Not Chain Succession (e, init)}
            \,\}
        \end{align*}
\begin{center}

    \digraph[scale=0.6]{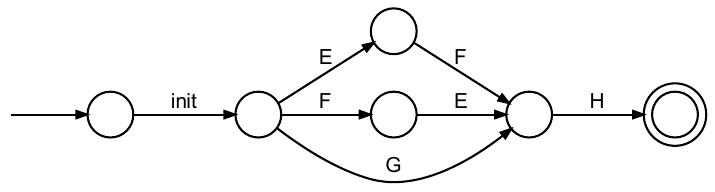}{margin="0.0,0.0"; rankdir="LR";
        edge [arrowsize="0.5"];
        start [label="",style = "invisible",width=0, height=0];
	node [height=".3",width=".3",shape="circle",label=""];
        1; 2; 3; 4; 5;
        6 [peripheries=2];
	start->1;
	1->2 [label="init",fontname="Arial",fontsize="10"];
	2->3 [label="E",fontname="Arial",fontsize="10"];
        3->5 [label="F",fontname="Arial",fontsize="10"];
        2->4 [label="F",fontname="Arial",fontsize="10"];
        4->5 [label="E",fontname="Arial",fontsize="10"];
        2->5 [label="G",fontname="Arial",fontsize="10"];
        5->6 [label="H",fontname="Arial",fontsize="10"];
        }%
\end{center}

\noindent
 \textsf{XOR-split in AND-split\,(\aname{A}, \aname{B\_or\_C}, \aname{D})} and 
 \textsf{XOR-join in AND-join\,(\aname{E\_or\_F}, \aname{G}, \aname{H})}. 
These are translated into the following Petri nets $\mathcal{N}_1$ and $\mathcal{N}_2$:
\begin{center}
    \digraph[scale=0.6]{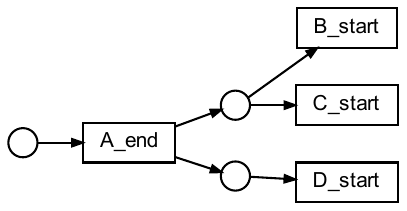}{ranksep=".3"; fontsize="10"; remincross=true; margin="0.0,0.0"; fontname="Arial"; rankdir="LR";
        edge [arrowsize="0.5"];
	node [height=".2",width=".2",fontname="Arial",fontsize="10"];
	"1" [shape="circle",label=""];
	"2" [shape="circle",label=""];
	"3" [shape="circle",label=""];
	"A_end" [shape="box"];
	"B_start" [shape="box"];
	"C_start" [shape="box"];
	"D_start" [shape="box"];
	"1" -> "A_end";
	"A_end" -> "2";
        "A_end" -> "3";
	"2" -> "B_start";
        "2" -> "C_start";
        "3" -> "D_start";}
    \digraph[scale=0.6]{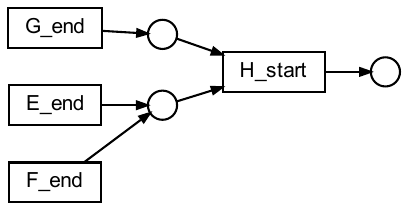}{ranksep=".3"; fontsize="10"; remincross=true; margin="0.0,0.0"; fontname="Arial"; rankdir="LR";
        edge [arrowsize="0.5"];
	node [height=".2",width=".2",fontname="Arial",fontsize="10"];
	"1" [shape="circle",label=""];
	"2" [shape="circle",label=""];
        "3" [shape="circle",label=""];
	"E_end" [shape="box"];
	"F_end" [shape="box"];
        "G_end" [shape="box"];
	"H_start" [shape="box"];
	"E_end" -> "1";
	"F_end" -> "1";
        "G_end" -> "2";
	"1" -> "H_start";
        "2" -> "H_start";
        "H_start" -> "3"}
\end{center}
These translations describe a similar situation to the two similar cases described above  
with the exception that the choice and parallel constructs are now swapped on Petri net graphs.
The DECLARE representation $\mathcal{D}_1$ of \textsf{XOR-split in an AND-split} can be rendered in the following way:
        \begin{align*}
            \mathcal{D}_1=\{\,&
                \textsf{Alternate Precedence(B, A)};
                \textsf{Alternate Precedence(C, A)};\\
                &\textsf{Alternate Succession(A, D)};
                \textsf{Alternate Succession(A, D)};\\
                &\textsf{Not CoExistence(B, C)};
                \textsf{Exactly One(A)}
            \,\}
        \end{align*}
\begin{center}
    \digraph[scale=0.6]{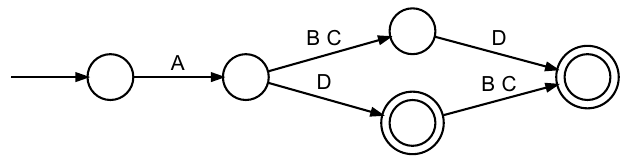}{margin="0.0,0.0"; rankdir="LR";
        edge [arrowsize="0.5"];
        start [label="",style = "invisible",width=0, height=0];
	node [height=".3",width=".3",shape="circle",label=""];
        1; 2; 3;
        4 [peripheries=2];
        5 [peripheries=2];
	start->1;
	1->2 [label="A",fontname="Arial",fontsize="10"];
	2->3 [label="B C",fontname="Arial",fontsize="10"];
        3->5 [label="D",fontname="Arial",fontsize="10"];
        2->4 [label="D",fontname="Arial",fontsize="10"];
        4->5 [label="B C",fontname="Arial",fontsize="10"];
        }
\end{center}
The DECLARE representation $\mathcal{D}_2$ of \textsf{XOR-join in an AND-join} is similarly obtained to the above one as follows: 
        \begin{align*}
            \mathcal{D}_2=\{\,&
                \textsf{Alternate Response(E, H)};
                \textsf{Alternate Response(F, H)};\\
                &\textsf{Alternate Succession(G, H)};
                \textsf{Not CoExistence(E, F)};\\
                &\textsf{Exactly One(H)}
            \,\}
        \end{align*}
\begin{center}
    \digraph[scale=0.6]{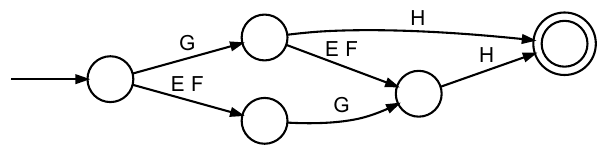}{margin="0.0,0.0"; rankdir="LR";
        edge [arrowsize="0.5"];
        start [label="",style = "invisible",width=0, height=0];
	node [height=".3",width=".3",shape="circle",label=""];
        1; 2; 3; 4;
        5 [peripheries=2];
	start->1;
	1->2 [label="G",fontname="Arial",fontsize="10"];
	1->3 [label="E F",fontname="Arial",fontsize="10"];
        2->4 [label="E F",fontname="Arial",fontsize="10"];
        3->4 [label="G",fontname="Arial",fontsize="10"];
        4->5 [label="H",fontname="Arial",fontsize="10"];
        2->5 [label="H",fontname="Arial",fontsize="10"];}%
\end{center}

\begin{figure*}
    \centering
    \includegraphics[width=\linewidth]{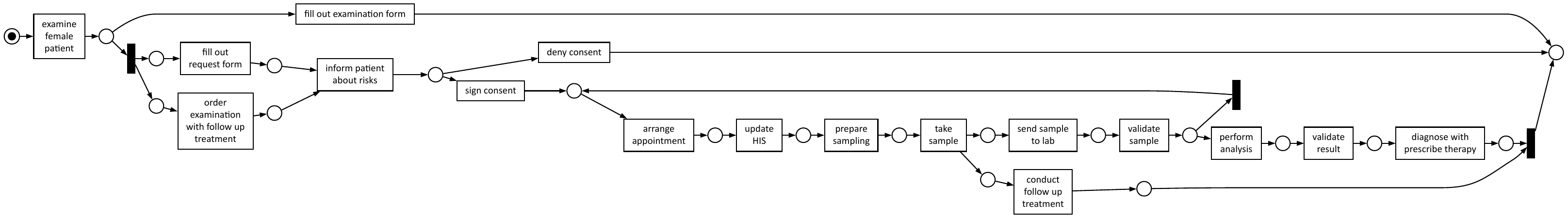}
    \caption{Petri net resulting from the automatic conversion of Fig.~\ref{lst:hospital-beepath}. Please note that the layout has been re-drawn for representation purposes and that ``start'' and ``end'' activities have been collapsed into a single transition.}
    \label{fig:hospital-petri}
\end{figure*}

\smallskip
Following the algorithm described in the beginning of the section, the final formal model is constructed 
by joining models obtained for each local fragment. In the case of DECLARE, the final model is the union of all local specifications. 
In the case of Petri nets, a slightly more elaborate treatment is necessary: local fragments are merged using Petri nets composition based on transition labels, i.e., the activity names. 
In Fig.~\ref{fig:hospital-petri}, we demonstrate the result of translating the text from Fig.~\ref{lst:hospital-beepath} into a Petri net.

\section{Tool Support}
\label{sec:tool}

A tool implementing the BeePath framework has been built as a web application using TypeScript and NodeJS.\footnote{Source code available at \url{https://github.com/dtu-pa/beepath} and deployment available at \url{https://dtu-pa.github.io/beepath/}.}

\begin{figure}[t]
    \centering
    \includegraphics[width=\linewidth]{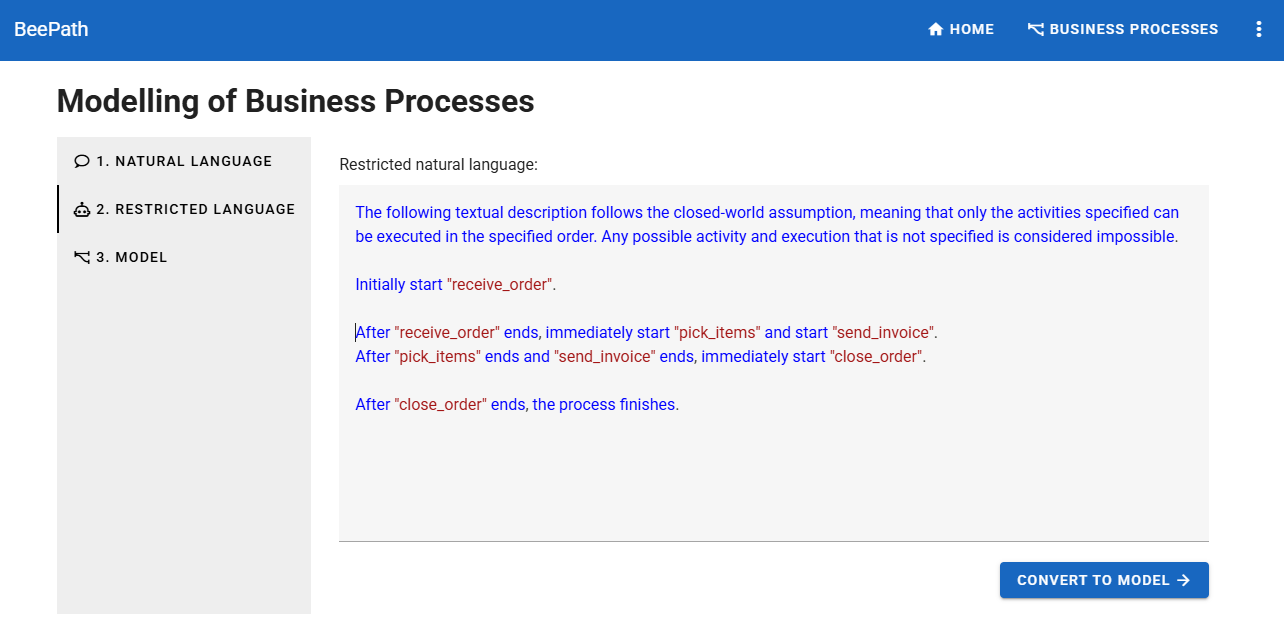}
    \caption{A screenshot of the implemented BeePath web application.}
    \label{fig:screenshot}
\end{figure}

The grammar for BeePath and its corresponding parsers were implemented using ANTLR,\footnote{See \url{https://www.antlr.org/}.} a parser generator that facilitates the automatic creation of parsers for structured languages based on a defined grammar. The output is rendered using Graphviz (for Petri nets and BPMN) and declare-js~\cite{nagel2023declare} (for DECLARE) and can also be downloaded as BPMN, TPN, or DECL files.

The connection to the LLM was established through a REST API request to the OpenAI platform\footnote{See \url{https://platform.openai.com/docs/overview}.} where all their models can be chosen and used (a valid key is required in the web application). The actual prompt being sent was refined through an iterative process and includes a general context overview (explaining that an unstructured text will be provided and must be converted into a structured format based on the given grammar), the EBNF grammar, and a set of conversion rules to address context-dependent scenarios. The conversion rules specify how to structure activity names (verb + noun), ensuring that every activity that starts must also have an endpoint, and require all subprocesses to be declared before being used.\footnote{The exact prompt is available at \url{https://github.com/dtu-pa/beepath/tree/main/src/converters/grammar/systemText.ts}.}

\bibliographystyle{splncs04}
\bibliography{references-mendeley,references}

\end{document}